\pdfoutput=1

\documentclass[11pt]{article}

\usepackage{acl}

\usepackage{times}
\usepackage{latexsym}
\usepackage{enumitem}
\usepackage{arydshln}

\usepackage[T1]{fontenc}

\usepackage[utf8]{inputenc}

\usepackage{microtype}
\usepackage{tabularx,ragged2e}
\usepackage{hyperref}
\usepackage{setspace}
\usepackage{caption}
\usepackage{multirow}
\usepackage{subcaption}
\usepackage{graphicx}
\usepackage{makecell}
\usepackage{pgfplots}
\usepackage{todonotes}
\usepackage{hyperref}
\usepackage{amsmath}
\usepackage{pifont}
\usepackage{latexsym}
\usepackage{graphicx}
\usepackage{booktabs}
\usepackage{float}
\usepackage{subcaption,booktabs}
\usepackage{times}
\usepackage{latexsym}
\usepackage{colortbl}
\usepackage{graphicx}
\usepackage{booktabs}
\usepackage{tcolorbox}
\usepackage{url}  

\usepackage{array}
\newcolumntype{L}[1]{>{\raggedright\let\newline\\\arraybackslash\hspace{0pt}}m{#1}}
\newcolumntype{C}[1]{>{\centering\let\newline\\\arraybackslash\hspace{0pt}}m{#1}}
\newcolumntype{R}[1]{>{\raggedleft\let\newline\\\arraybackslash\hspace{0pt}}m{#1}}
\usepackage{enumitem}
\definecolor{OliveGreen}{rgb}{0,0.6,0}

\usepackage{microtype}

\title{Show, Don't Tell: Demonstrations Outperform Descriptions for Schema-Guided Task-Oriented Dialogue}

\author{
    Raghav Gupta\thanks{*Equal contribution}, 
    Harrison Lee\footnotemark[1], 
    Jeffrey Zhao, 
    Abhinav Rastogi, 
    Yuan Cao, 
    Yonghui Wu\\
Google Research \\
\texttt{\{raghavgupta, harrisonlee\}@google.com}}

\begin{document}
\maketitle
\begin{abstract}
Building universal dialogue systems that operate across multiple domains/APIs and generalize to new ones with minimal overhead is a critical challenge. Recent works have leveraged natural language descriptions of schema elements to enable such systems; however, descriptions only indirectly convey schema semantics. In this work, we propose \textit{Show, Don't Tell}, which prompts seq2seq models with a labeled example dialogue to \textit{show} the semantics of schema elements rather than \textit{tell} the model through descriptions. While requiring similar effort from service developers as generating descriptions, we show that using short examples as schema representations with large language models results in state-of-the-art performance on two popular dialogue state tracking benchmarks designed to measure zero-shot generalization - the Schema-Guided Dialogue dataset and the MultiWOZ leave-one-out benchmark.
\end{abstract}

\section{Introduction}
Task-oriented dialogue (TOD) systems need to support an ever-increasing variety of services. Since many service developers lack the resources to collect data and train models, zero and few-shot transfer to unseen services is critical to the democratization of dialogue agents. 

Recent approaches to generalizable TOD systems primarily rely on combining two techniques: large language models like BERT \citep{bert} and T5 \citep{raffel2020exploring}, and schema-guided modeling - i.e. using natural language descriptions of schema elements (intents and slots) as model inputs to enable transfer to unseen services \citep{rastogi2020schema,rastogi2020towards}. Models combining the two currently hold state-of-the-art (SotA) results on dialogue state tracking (DST) \citep{heck-etal-2020-trippy, lee2021dialogue, zhao2022description}.

\begin{figure*}[t!]
\centering
\begin{minipage}{15.6cm}\vspace{0mm}    \centering
\begin{tcolorbox}[colback=yellow!5!white]
    \centering
    \small
    \begin{tabular}{ L {57mm}  L {77mm} }
    \arrayrulecolor{Gray}
    \textbf{T5-ind} & \textbf{SDT-ind} \\
    $\mathtt{P_1}$ = amount: The amount of money to send or request & $\mathtt{P^{ind}_1}$ = [ex] \textcolor{blue}{[user] I need to transfer 125 dollars} \textcolor{OliveGreen}{[slot] amount=125 dollars}\\ \arrayrulecolor{black} $\mathtt{P_2}$ = receiver: Name of the contact or account to make the transaction with & $\mathtt{P^{ind}_2}$ = [ex] \textcolor{blue}{[user] Make the transfer to Victoria.} \textcolor{OliveGreen}{[slot] receiver=Victoria}\\ 
    \ldots & \ldots\\
    \midrule
    \textbf{T5-seq} & \textbf{SDT-seq}  \\
    
$\mathtt{P}$ = 0: The amount of money to send or request 1: Name of the contact or account to make the transaction with 2: Whether the transaction is private or not a) True b) False 3: The source of money used for making the payment a) credit card b) debit card c) app balance  & $\mathtt{P^{seq}}$ = [ex] \textcolor{blue}{[user] I want to make a payment to Jerry for \$82 from my mastercard} 
\textcolor{red}{[system] Confirming you want to pay Jerry \$82 with your credit card yes?} 
\textcolor{blue}{[user] Yes that's right, make the transaction private too}
\textcolor{OliveGreen}{[slot] amount=\$82 receiver=Jerry private\_visibility=a of a) True b) False payment\_method=a of a) credit card b) debit card c) app balance}
    \end{tabular}
\end{tcolorbox}
\vspace{-2mm}
\caption{Illustration of all prompt formats for a payment service for both description-based and \textit{Show, Don't Tell} models with independent (top) and sequential (bottom) decoding of dialogue state.}
    \label{fig:sdt_example}
\end{minipage}
\end{figure*}

However, description-based schema representations have some drawbacks. Writing precise natural language descriptions requires manual effort and can be difficult to write succinctly. Also, descriptions only provide indirect supervision about how to interact with a service compared to an example. Furthermore, \citet{Lee_Gupta_Rastogi_Cao_Zhang_Wu_2022} showed that schema-guided DST models are not robust to variations in schema descriptions, causing significant quality drops.

We propose using a single dialogue example with state annotations as an alternative to the description-based schema representation, similar to one-shot priming \citep{brown2020language} - an approach we call \textit{Show, Don't Tell (SDT)}. Through demonstration, we \textbf{show} models the schema semantics rather than \textbf{tell} them through natural language descriptions, as seen in Figure \ref{fig:sdt_example}. SDT achieves SotA accuracy and generalization to new APIs across both the Schema-Guided Dataset (SGD) \citep{rastogi2020towards} and MultiWOZ Leave-One-Out \citep{budzianowski-etal-2018-multiwoz, lin2021leveraging} benchmarks, while being more data-efficient and robust to schema variations.

\section{Show, Don't Tell}
\label{sec:methodology}

Following SoTA models, we pose DST as a seq2seq task \citep{wu2019transferable, zhao2021effective} and finetune T5 on DST datasets. The model input consists of a \textit{prompt} to convey API semantics and \textit{context} to represent the current dialogue instance. The \textit{target} contains ground truth belief states corresponding to the context. We compare against two baselines:

\begin{itemize}[leftmargin=*]
    \item \textbf{T5-ind} \cite{lee2021dialogue}: Model input comprises \textit{a single slot description} for the prompt, concatenated with the dialogue history as the context. The target is the value of the single slot in the dialogue state. Model inference is invoked once per slot - i.e. values for different slots are independently decoded.
    \item \textbf{T5-seq} \cite{zhao2022description}: Model input comprises the descriptions of \textit{all slots} as the prompt, concatenated with the dialogue history as the context. The target is the sequence of slot-value pairs in the dialogue state - i.e. the dialogue state is decoded sequentially in a single pass.
\end{itemize}

We modify the prompt formats above to utilize demonstrations instead of descriptions as described below and illustrated in Figure \ref{fig:sdt_example}.

\begin{itemize}[leftmargin=*]
    \item \textbf{SDT-ind}: A \textit{prompt} $\mathtt{P^{ind}_i}$ comprises a single example utterance and the ground truth slot-value pair formatted as
    \begin{equation*}
\mathtt{P^{ind}_{i} = [ex]; u^{ind}_i; [slot]; sv_i}
\end{equation*}
\noindent where $\mathtt{u^{ind}_i}$ is a user utterance where slot $\mathtt{i}$ is active/not null and $\mathtt{sv_i}$ is the slot-value pair. $\mathtt{[ex], [slot]}$ are special delimiter tokens, and $;$ denotes concatenation.
    \item \textbf{SDT-seq}: A \textit{prompt} $\mathtt{P^{seq}}$ comprises a single labeled dialogue formatted as:
\begin{equation*}
\mathtt{P^{seq} = [ex]; u_{1}; ...; u_{n}; [slot]; sv_{1}; ...; sv_{m}}
\end{equation*}

where $\mathtt{u_j}$ is an utterance, and other symbols are explained in the SDT-ind section above. In simple terms, the prompt is constructed by concatenating all utterances in an example dialogue followed by all slot-value pairs in the dialogue state.
\end{itemize}

In both the T5-* and SDT-* approaches, the context is the serialized dialogue history for the current dialogue instance. The final model input is formed by concatenating the prompt and the context strings, and the target string is the same as T5-*, containing only a single slot value for *-ind models and the entire turn's belief state for *-seq models.

For both T5-* and SDT-*, we enumerate the categorical slot values in multiple-choice format in the prompt and task models with decoding the multiple choice letter corresponding to the correct categorical value.

More details on prompt design and its impact on performance are provided in Appendix \ref{sec:prompt_design_study}.

\textbf{Creating prompt examples:} It is imperative that SDT prompts contain enough information to infer the semantics for all slots in a schema. For SDT-ind, we create individual utterances that showcase a single slot. For SDT-seq, we create example dialogues where all slots in the schema are used. 

\textbf{Multi-domain examples:} It is not feasible to construct multi-domain demonstrations for every combination of domains. Thus, we stick to single-domain SDT prompts and create separate training instances for each domain present in a multi-domain dialogue turn; for inference, we run inference once for each domain and combine the results.

\section{Experimental Setup}

\textbf{Datasets:} We conduct experiments on two DST benchmarks: Schema-guided Dialogue (SGD) \citep{rastogi2020towards} and MultiWOZ 2.1 \citep{budzianowski-etal-2018-multiwoz,eric-etal-2020-multiwoz}. For MultiWOZ, we evaluate on the leave-one-out setup \citep{wu2019transferable,lin2021zeroshot}, where models are trained on all domains but one and evaluated on the holdout domain. Additionally, we apply the recommended TRADE pre-processing script\footnote{\url{https://github.com/budzianowski/multiwoz##dialog-state-tracking}} for fair comparison with other work. For both datasets, we created concise example dialogues modeled after dialogues observed in the datasets.

\textbf{Implementation:} We train SDT models by fine-tuning pretrained T5 1.1 checkpoints. For SDT-seq, we select one example dialogue for each service to create a prompt and use that prompt across all dialogue instances of that service, across training and evaluation. We do the same for SDT-ind but create one prompt per slot instead of per service. Unless otherwise noted, all T5-based models are based on T5-XXL (11B parameters). Appendices \ref{sec:sdt_model_details} and \ref{sec:baseline_models} contain more details on training and baselines respectively.

\section{Results}

\subsection{SGD Results}

Table \ref{table:sdt_vs_others} contains results on the SGD test set. SDT-seq achieves the highest JGA by +1.1\%, outperforming the description-based T5-* models, particularly on unseen services. SDT-ind is comparable to its counterpart T5-ind and better than T5-seq.

Since SDT results vary with the choice of example dialogue provided in the prompt, we created 5 different versions of prompts for each service using different examples. We report the average JGA across the 5 versions and the 95\% confidence intervals using the Student's-t distribution. 

We hypothesize that the main advantage of SDT is that the schema semantics are conveyed via demonstration, which is more similar in form to the end task of state tracking and more informative than descriptions. On the other hand, natural language descriptions can be viewed as an intermediary that models must interpret in order to achieve the end goal of slot value prediction.

We see that SDT-seq outperforms SDT-ind and posit that this is because the full dialogue prompts in SDT-seq demonstrate more complex linguistic patterns (e.g. coreference resolution, long term dependencies) than the single utterance prompts of SDT-ind. On the other hand, we believe T5-seq does not outperform T5-ind because no additional information is conveyed to the model through concatenating independent descriptions. All-else-equal, decoding all slots in one pass is more challenging than decoding each slot independently.

We also experimented with using up to 5 example dialogues in each prompt of SDT-seq, but accuracy did not increase.

\begin{table}[t!]
\centering
\setlength{\tabcolsep}{4pt}
\scalebox{0.9}{
    \begin{tabular}{p{2.8cm}ccc}
    \hline
    \textbf{Model} & All & Seen & Unseen \\
    \hline
    MRC+WD-DST* & 86.5 & 92.4 & 84.6 \\
    T5-seq & 86.4 & \textbf{95.8} & 83.3 \\
    T5-ind & 87.7 & 95.3 & 85.2 \\
    \hline
    SDT-ind & 87.5$\pm$0.9 & 95.2$\pm$0.7 & 85.0$\pm$1.4 \\
    SDT-seq & \textbf{88.8}$\pm$\textbf{0.5} & \textbf{95.8}$\pm$\textbf{0.2} & \textbf{86.4}$\pm$\textbf{0.7} \\
    \hline
    \end{tabular}
}
\caption{SDT achieves state-of-the-art JGA as evaluated on the SGD test set, performing especially well on unseen services. *Data augmentation/special rules applied.}
\label{table:sdt_vs_others}
\end{table}

\begin{table}[t!]
\centering
\setlength{\tabcolsep}{1.5pt}
    \scalebox{0.85}{
    \begin{tabular}{l|ccccc|c}
    \hline
    \textbf{Model} & Attraction & Hotel & Restaurant & Taxi & Train & Avg \\
    \hline
    TRADE & 20.1 & 14.2 & 12.6 & 59.2 & 22.4 & 25.7 \\
    SUMBT & 22.6 & 19.8 & 16.5 & 59.5 & 22.5 & 28.2 \\
    TransferQA & 31.3 & 22.7 & 26.3 & 61.9 & 36.7 & 35.8 \\
    T5-seq & \textbf{76.1} & 28.6 & 69.8 & \textbf{87.0} & 60.4 & 64.4 \\
    \hline
    SDT-seq & 74.4 & \textbf{33.9} & \textbf{72.0} & 86.4 & \textbf{62.9} & \textbf{65.9} \\
    \hline
    \end{tabular}
    }
\caption{SDT-seq outperforms T5-seq on the MultiWOZ 2.1 cross-domain (leave-one-out) benchmark. Results for TRADE, SUMBT, and TransferQA from \citet{kumar2020ma}, \citet{campagna-etal-2020-zero}, and \citet{lin2021zeroshot}, respectively.}
\label{table:mw_cross_domain}
\end{table}

\subsection{MultiWOZ Results}
Table \ref{table:mw_cross_domain} summarizes results for the MultiWOZ 2.1 leave-one-out setup. SDT-seq outperforms T5-seq by +1.5\% overall and in 3 of the 5 domains, achieving state-of-the-art performance. 

\subsection{Impact of Model Size}

T5's XXL size (11B parameters) may be unsuitable in resource-constrained settings. To understand how the the impact of model size, we measure SDT's performance on SGD across multiple T5 sizes in Table \ref{table:sdt_size}. For base and large sizes, both SDT variations offer higher JGA than their description-based counterparts, possibly due to smaller T5 models being less capable of inferring unseen slots with just a description, whereas SDT models provide more direct supervision in contrast. Additionally, SDT-ind outperforms SDT-seq for both the smaller sizes, potentially due to SDT-seq's prediction task being more complex than that of SDT-ind.

\begin{table}[h!]
\centering
\setlength{\tabcolsep}{2pt}
\scalebox{0.95}{
    \begin{tabular}{p{1.4cm}ccc}
    \hline
    \textbf{Model} & Base (250M) & Large (800M) & XXL (11B) \\
    \hline
    T5-seq & 72.9 & 80.0 & 86.4 \\
    T5-ind & 72.6 & 82.2 & 87.7 \\
    \hline
    SDT-ind & \textbf{78.2}$\pm$\textbf{0.6} & \textbf{83.7}$\pm$\textbf{0.8} & 87.5$\pm$0.9 \\
    SDT-seq & 76.3$\pm$1.6 & 83.2$\pm$0.6 & \textbf{88.8}$\pm$\textbf{0.5} \\
    \hline
    \end{tabular}
}
\caption{SGD test set JGA across T5's Base, Large, and XXL sizes. SDT's advantage is especially prominent on smaller model sizes.}
\label{table:sdt_size}
\end{table}

\subsection{Data Efficiency}

To examine the data efficiency of SDT models, we also experiment with training SDT-seq with 0.16\% (10-shot), 1\%, and 10\% of the SGD training data and evaluating on the entire test set. For 10-shot, we randomly sample 10 training dialogues from every service; for 1\% and 10\%, we sample uniformly across the entire dataset. SDT-seq demonstrates far higher data efficiency than T5-seq (Table \ref{table:data_efficiency}), indicating that SDT is more suitable for bootstrapping dialogue systems with a limited budget for collecting training data.

\begin{table}[ht!]
\centering
\setlength{\tabcolsep}{3.4pt}
\scalebox{0.95}{
    \begin{tabular}{p{2cm}ccc}
    \hline
    \textbf{Model}    & 10-shot & 1\% & 10\% \\
    \hline
    T5-seq  & 51.0 
    & 79.4 
    & 83.0 
    \\
    SDT-seq & \textbf{70.7} & \textbf{84.5} & \textbf{87.4} \\
    \hline
    \end{tabular}
}
\caption{Data efficiency experiments on the SGD test set. SDT-seq's example-based prompt approach is more suited to low resource settings than T5-seq's description-based prompts.}
\label{table:data_efficiency}
\end{table}

\subsection{Robustness}

Large LMs are often sensitive to the choice of prompt \citep{zhao2021calibrate,reynolds2021prompt}. To this end, we evaluate SDT-seq on the SGD-X \citep{Lee_Gupta_Rastogi_Cao_Zhang_Wu_2022} benchmark, comprising 5 variants with paraphrased slot names and descriptions for every schema (Appendix Figure \ref{fig:sgdx_schemas}). Note that SDT-seq only makes use of slot names, so variations in description have no effect on it.

Table \ref{table:sgdx} shows SDT-seq achieves the highest average JGA ($JGA_{v_{1-5}}$) and lowest schema sensitivity ($SS_{JGA}$, lower value indicates higher robustness), making it the most robust of the compared models. While the JGA decline indicates that SDT-seq is somewhat sensitive to how slot names are written, when compared to a variant of T5-seq \citep{zhao2022description} that only uses slot names, it is still more robust based on the schema sensitivity, and the relative drop in JGA is nearly equal.

\begin{table}[ht!]

\centering
\setlength\tabcolsep{3.4pt}
    \scalebox{0.84}{
    \begin{tabular}{lcccc}
    \hline
    \textbf{Model} & $JGA_{Orig}$ & $JGA_{v_{1-5}}$ & $Diff_{rel}$ & $SS_{JGA}$\\
    \hline
    SGP-DST* & 60.5 & 49.9 & -17.5 & 51.9 \\
    T5-ind$_{base}$* & 72.6 & 64.0 & -11.9 & 40.4 \\
    T5-seq (name)\ding{72} & 79.7 & 73.0 & \textbf{-8.4} & 35.0 \\
    T5-seq & 86.4 & 77.8 & -10.0 & 27.0 \\
    \hline
    SDT-seq & \textbf{88.8} & \textbf{81.2} & -8.6 & \textbf{24.1} \\
    \hline
    \end{tabular}
    }
\caption{Robustness evaluation on the SGD-X test sets. *Results from \citet{Lee_Gupta_Rastogi_Cao_Zhang_Wu_2022}. \ding{72}Result of using T5-seq with only slot names and no descriptions, from \citet{zhao2022description}.}
\label{table:sgdx}
\end{table}



\begin{figure*}[t!]
\centering
\begin{minipage}{15.6cm}\vspace{0mm}    \centering
\begin{tcolorbox}[colback=blue!5]
    \centering
    \small
    \begin{tabular}{ L {75mm} L {60mm} }
     \textbf{Example Dialogue}& \textbf{Predictions} \\
    \midrule
    \textbf{1. Disambiguating similar slots} & \multirow{2}{45mm}{T5-seq: \textcolor{red}{\textit{to=Sacramento},  \textit{from=Anaheim}} \\ SDT-seq: \textcolor{blue}{\textit{to=Anaheim},  \textit{from=Sacramento}}}\\
        
        [user] I need to find tickets to \textbf{Anaheim}, CA. [system] When would you like to travel, and where are you going to? [user] Traveling to \textbf{Sacramento} on the 4th.
    &  \\\\ 
        \textbf{2. Handling unseen slots} & \multirow{2}{55mm}{T5-seq: \textcolor{red}{\textit{new\_alarm\_name=None}} \\ 
        SDT-seq: \textcolor{blue}{\textit{new\_alarm\_name=Grocery run}}}\\
        
        [user] Can you please add an alarm called Grocery run. \\
        
        \midrule
        
        \textbf{3. Predicting categorical values not seen in prompt} & \multirow{2}{45mm}{T5-seq: \textcolor{blue}{\textit{event\_type=theater}} \\
        SDT-seq: \textcolor{red}{\textit{event\_type=music}}}\\
        
        [user] I like Broadway shows and want to see one on Tuesday next week.
    \end{tabular}
\end{tcolorbox}
\caption{Comparing common error patterns made by T5-seq vs. SDT-seq. Correct and incorrect predictions colored in red and blue, respectively.}
    \label{fig:error}
\end{minipage}
\end{figure*}

\section{Discussion}
\label{sec:discussion}

\subsection{Writing descriptions vs. demonstrations}

The information provided to SDT is not identical to what is provided to typical schema-guided models, as SDT exchanges natural language descriptions for a demonstration of identifying slots in a dialogue. However, we argue that from the developer standpoint, creating a single example is similar in effort to writing descriptions, so we consider the methods comparable. Creating the SDT-seq prompts for all 45 services in SGD took an experienced annotator $\sim$2 hours, compared to $\sim$1.5 hours for generating all slot descriptions. SDT-ind prompts are even simpler to write because they relax the requirement for creating a coherent dialogue involving all slots.

Descriptions can sometimes be easier to generate than a succinct dialogue that covers all slots. However, given the performance gain, example-based prompts may be a better choice for many settings, especially for smaller model sizes and low resource settings where the gain over description-based prompts is more pronounced.

\subsection{Descriptions plus demonstrations}
We tried combining both descriptions and a demonstration in a single prompt to try to further improve performance. However, results showed that this did not improve upon using demonstrations alone (see Appendix Table \ref{table:desc_and_demo} for details). 

We hypothesize that demonstrations, along with slot names, already convey slot semantics sufficiently, rendering descriptions extraneous. However, given that using slot names alone underperforms using descriptions  \citep{zhao2022description}, the improvement SDT exhibits over using descriptions does not result purely from the use of slot names.

\subsection{Prompting vs. traditional finetuning}

To understand the impact of using a single demonstration as a prompt vs. traditional finetuning, we finetune T5-seq an additional time on the same set of dialogues used in SDT-seq prompts; therefore it has access to both slot descriptions as well as a single demonstration for each service. In this case, T5-seq is provided strictly more information than SDT-seq. T5-seq with finetuning obtains a JGA of 87.7\% on SGD, on par with T5-ind but still lower than SDT-seq, suggesting that, when scarce, dialogue examples are better used as prompts \citep{le2021many}. 

Interestingly, finetuning on up to 5 dialogue examples per service did not improve performance after the first example (Appendix Figure \ref{fig:t5seq_finetune}).

\subsection{Error analysis}
Figure \ref{fig:error} compares some common error patterns made by T5-seq vs. SDT-seq. The patterns suggest that SDT's demonstrations are helpful when multiple slots in the same domain are similar to each other (\#1 in Figure \ref{fig:error}) and when slots dissimilar from those seen in training are introduced (\#2). However, SDT can sometimes be limited by its prompt. For instance, in \#3 it has only seen the "music" value for the \textit{event\_type} slot in the prompt, potentially resulting in under-predicting the categorical values not featured in the example dialogue (e.g. "theater").


\section{Related Work}

Prior approaches focused on framing DST as question answering \citep{ruan2020fine,ma2019end,zhang2021sgdqa}. Many MultiWOZ cross-domain models leverage slot names/descriptions \citep{wu2019transferable,lee2019sumbt,lin2021zeroshot}. 

Pretrained generative LLMs \citep{raffel2020exploring, brown2020language} have enabled framing NLP tasks as seq2seq problems. Some DST papers \citep{zhao2021effective,feng-etal-2021-sequence} look at settings with no train-test discrepancy. Many studies explore the efficacy of task-specific prompts \citep{jiang2020know,liu2021gpt}. \citet{madotto2020language} and prime LMs with examples for dialogue tasks, but without finetuning. \citet{wei2021finetuned} finetunes language models to teach them to use prompts to generalize across NLP tasks.

\section{Conclusion}

We study the use of demonstrations as LM prompts to convey the semantics of APIs in lieu of natural language descriptions for TOD. While taking similar effort to construct, demonstrations outperform description-based prompts in our experiments across DST datasets (SGD and MultiWOZ), model sizes, and training data sizes, while being more robust to changes in schemata. This work provides developers of TOD systems with more options for API representations to enable transfer to unseen services. In future work, we would like to explore this representation for other TOD tasks (e.g. dialogue management and response generation). 

\section{Ethical Considerations}
We proposed a more efficient way of building TOD systems by leveraging demonstrations in place of descriptions, leading to increased accuracy with minimal/no data preparation overhead. We conduct our experiments on publicly-available TOD datasets in English, covering domains which are popular for building conversational agents. We hope our work leads to building more accurate TOD systems with similar or less overhead and encourages further research in the area.


\bibliography{anthology,custom}

\begin{thebibliography}{30}
\expandafter\ifx\csname natexlab\endcsname\relax\def\natexlab#1{#1}\fi

\bibitem[{Brown et~al.(2020)Brown, Mann, Ryder, Subbiah, Kaplan, Dhariwal,
  Neelakantan, Shyam, Sastry, Askell, Agarwal, Herbert-Voss, Krueger, Henighan,
  Child, Ramesh, Ziegler, Wu, Winter, Hesse, Chen, Sigler, Litwin, Gray, Chess,
  Clark, Berner, McCandlish, Radford, Sutskever, and
  Amodei}]{brown2020language}
Tom Brown, Benjamin Mann, Nick Ryder, Melanie Subbiah, Jared~D Kaplan, Prafulla
  Dhariwal, Arvind Neelakantan, Pranav Shyam, Girish Sastry, Amanda Askell,
  Sandhini Agarwal, Ariel Herbert-Voss, Gretchen Krueger, Tom Henighan, Rewon
  Child, Aditya Ramesh, Daniel Ziegler, Jeffrey Wu, Clemens Winter, Chris
  Hesse, Mark Chen, Eric Sigler, Mateusz Litwin, Scott Gray, Benjamin Chess,
  Jack Clark, Christopher Berner, Sam McCandlish, Alec Radford, Ilya Sutskever,
  and Dario Amodei. 2020.
\newblock \href
  {https://proceedings.neurips.cc/paper/2020/file/1457c0d6bfcb4967418bfb8ac142f64a-Paper.pdf}
  {Language models are few-shot learners}.
\newblock In \emph{Advances in Neural Information Processing Systems},
  volume~33, pages 1877--1901. Curran Associates, Inc.

\bibitem[{Budzianowski et~al.(2018)Budzianowski, Wen, Tseng, Casanueva, Ultes,
  Ramadan, and Ga{\v{s}}i{\'c}}]{budzianowski-etal-2018-multiwoz}
Pawe{\l} Budzianowski, Tsung-Hsien Wen, Bo-Hsiang Tseng, I{\~n}igo Casanueva,
  Stefan Ultes, Osman Ramadan, and Milica Ga{\v{s}}i{\'c}. 2018.
\newblock \href {https://doi.org/10.18653/v1/D18-1547} {{M}ulti{WOZ} - a
  large-scale multi-domain {W}izard-of-{O}z dataset for task-oriented dialogue
  modelling}.
\newblock In \emph{Proceedings of the 2018 Conference on Empirical Methods in
  Natural Language Processing}, pages 5016--5026, Brussels, Belgium.
  Association for Computational Linguistics.

\bibitem[{Campagna et~al.(2020)Campagna, Foryciarz, Moradshahi, and
  Lam}]{campagna-etal-2020-zero}
Giovanni Campagna, Agata Foryciarz, Mehrad Moradshahi, and Monica Lam. 2020.
\newblock \href {https://doi.org/10.18653/v1/2020.acl-main.12} {Zero-shot
  transfer learning with synthesized data for multi-domain dialogue state
  tracking}.
\newblock In \emph{Proceedings of the 58th Annual Meeting of the Association
  for Computational Linguistics}, pages 122--132, Online. Association for
  Computational Linguistics.

\bibitem[{Devlin et~al.(2019)Devlin, Chang, Lee, and Toutanova}]{bert}
Jacob Devlin, Ming-Wei Chang, Kenton Lee, and Kristina Toutanova. 2019.
\newblock \href {https://doi.org/10.18653/v1/N19-1423} {{BERT}: Pre-training of
  deep bidirectional transformers for language understanding}.
\newblock In \emph{Proceedings of the 2019 Conference of the North {A}merican
  Chapter of the Association for Computational Linguistics: Human Language
  Technologies, Volume 1 (Long and Short Papers)}, pages 4171--4186,
  Minneapolis, Minnesota. Association for Computational Linguistics.

\bibitem[{Eric et~al.(2020)Eric, Goel, Paul, Sethi, Agarwal, Gao, Kumar, Goyal,
  Ku, and Hakkani-Tur}]{eric-etal-2020-multiwoz}
Mihail Eric, Rahul Goel, Shachi Paul, Abhishek Sethi, Sanchit Agarwal, Shuyang
  Gao, Adarsh Kumar, Anuj Goyal, Peter Ku, and Dilek Hakkani-Tur. 2020.
\newblock \href {https://aclanthology.org/2020.lrec-1.53} {{M}ulti{WOZ} 2.1: A
  consolidated multi-domain dialogue dataset with state corrections and state
  tracking baselines}.
\newblock In \emph{Proceedings of the 12th Language Resources and Evaluation
  Conference}, pages 422--428, Marseille, France. European Language Resources
  Association.

\bibitem[{Feng et~al.(2021)Feng, Wang, and Li}]{feng-etal-2021-sequence}
Yue Feng, Yang Wang, and Hang Li. 2021.
\newblock \href {https://doi.org/10.18653/v1/2021.acl-long.135} {A
  sequence-to-sequence approach to dialogue state tracking}.
\newblock In \emph{Proceedings of the 59th Annual Meeting of the Association
  for Computational Linguistics and the 11th International Joint Conference on
  Natural Language Processing (Volume 1: Long Papers)}, pages 1714--1725,
  Online. Association for Computational Linguistics.

\bibitem[{Heck et~al.(2020)Heck, van Niekerk, Lubis, Geishauser, Lin, Moresi,
  and Gasic}]{heck-etal-2020-trippy}
Michael Heck, Carel van Niekerk, Nurul Lubis, Christian Geishauser, Hsien-Chin
  Lin, Marco Moresi, and Milica Gasic. 2020.
\newblock \href {https://aclanthology.org/2020.sigdial-1.4} {{T}rip{P}y: A
  triple copy strategy for value independent neural dialog state tracking}.
\newblock In \emph{Proceedings of the 21th Annual Meeting of the Special
  Interest Group on Discourse and Dialogue}, pages 35--44, 1st virtual meeting.
  Association for Computational Linguistics.

\bibitem[{Jiang et~al.(2020)Jiang, Xu, Araki, and Neubig}]{jiang2020know}
Zhengbao Jiang, Frank~F. Xu, Jun Araki, and Graham Neubig. 2020.
\newblock \href {http://arxiv.org/abs/1911.12543} {How can we know what
  language models know?}

\bibitem[{Jouppi et~al.(2017)Jouppi, Young, Patil, Patterson, Agrawal, Bajwa,
  Bates, Bhatia, Boden, Borchers, Boyle, and Cantin}]{tpu2017}
Norman~P. Jouppi, Cliff Young, Nishant Patil, David Patterson, Gaurav Agrawal,
  Raminder Bajwa, Sarah Bates, Suresh Bhatia, Nan Boden, Al~Borchers, Rick
  Boyle, and Pierre-luc et~al. Cantin. 2017.
\newblock \href {https://doi.org/10.1145/3140659.3080246} {In-datacenter
  performance analysis of a tensor processing unit}.
\newblock \emph{SIGARCH Comput. Archit. News}, 45(2):1–12.

\bibitem[{Kumar et~al.(2020)Kumar, Ku, Goyal, Metallinou, and
  Hakkani-Tur}]{kumar2020ma}
Adarsh Kumar, Peter Ku, Anuj Goyal, Angeliki Metallinou, and Dilek Hakkani-Tur.
  2020.
\newblock \href {https://doi.org/10.1609/aaai.v34i05.6322} {Ma-dst:
  Multi-attention-based scalable dialog state tracking}.
\newblock \emph{Proceedings of the AAAI Conference on Artificial Intelligence},
  34(05):8107--8114.

\bibitem[{Le~Scao and Rush(2021)}]{le2021many}
Teven Le~Scao and Alexander Rush. 2021.
\newblock \href {https://doi.org/10.18653/v1/2021.naacl-main.208} {How many
  data points is a prompt worth?}
\newblock In \emph{Proceedings of the 2021 Conference of the North American
  Chapter of the Association for Computational Linguistics: Human Language
  Technologies}, pages 2627--2636, Online. Association for Computational
  Linguistics.

\bibitem[{Lee et~al.(2021)Lee, Cheng, and Ostendorf}]{lee2021dialogue}
Chia-Hsuan Lee, Hao Cheng, and Mari Ostendorf. 2021.
\newblock \href {https://doi.org/10.18653/v1/2021.emnlp-main.404} {Dialogue
  state tracking with a language model using schema-driven prompting}.
\newblock In \emph{Proceedings of the 2021 Conference on Empirical Methods in
  Natural Language Processing}, pages 4937--4949, Online and Punta Cana,
  Dominican Republic. Association for Computational Linguistics.

\bibitem[{Lee et~al.(2022)Lee, Gupta, Rastogi, Cao, Zhang, and
  Wu}]{Lee_Gupta_Rastogi_Cao_Zhang_Wu_2022}
Harrison Lee, Raghav Gupta, Abhinav Rastogi, Yuan Cao, Bin Zhang, and Yonghui
  Wu. 2022.
\newblock \href {https://doi.org/10.1609/aaai.v36i10.21341} {Sgd-x: A benchmark
  for robust generalization in schema-guided dialogue systems}.
\newblock \emph{Proceedings of the AAAI Conference on Artificial Intelligence},
  36(10):10938--10946.

\bibitem[{Lee et~al.(2019)Lee, Lee, and Kim}]{lee2019sumbt}
Hwaran Lee, Jinsik Lee, and Tae-Yoon Kim. 2019.
\newblock \href {https://doi.org/10.18653/v1/P19-1546} {{SUMBT}: Slot-utterance
  matching for universal and scalable belief tracking}.
\newblock In \emph{Proceedings of the 57th Annual Meeting of the Association
  for Computational Linguistics}, pages 5478--5483, Florence, Italy.
  Association for Computational Linguistics.

\bibitem[{Lin et~al.(2021{\natexlab{a}})Lin, Liu, Madotto, Moon, Crook, Zhou,
  Wang, Yu, Cho, Subba, and Fung}]{lin2021zeroshot}
Zhaojiang Lin, Bing Liu, Andrea Madotto, Seungwhan Moon, Paul Crook, Zhenpeng
  Zhou, Zhiguang Wang, Zhou Yu, Eunjoon Cho, Rajen Subba, and Pascale Fung.
  2021{\natexlab{a}}.
\newblock \href {http://arxiv.org/abs/2109.04655} {Zero-shot dialogue state
  tracking via cross-task transfer}.

\bibitem[{Lin et~al.(2021{\natexlab{b}})Lin, Liu, Moon, Crook, Zhou, Wang, Yu,
  Madotto, Cho, and Subba}]{lin2021leveraging}
Zhaojiang Lin, Bing Liu, Seungwhan Moon, Paul Crook, Zhenpeng Zhou, Zhiguang
  Wang, Zhou Yu, Andrea Madotto, Eunjoon Cho, and Rajen Subba.
  2021{\natexlab{b}}.
\newblock \href {https://doi.org/10.18653/v1/2021.naacl-main.448} {Leveraging
  slot descriptions for zero-shot cross-domain dialogue {S}tate{T}racking}.
\newblock In \emph{Proceedings of the 2021 Conference of the North American
  Chapter of the Association for Computational Linguistics: Human Language
  Technologies}, pages 5640--5648, Online. Association for Computational
  Linguistics.

\bibitem[{Liu et~al.(2021)Liu, Zheng, Du, Ding, Qian, Yang, and
  Tang}]{liu2021gpt}
Xiao Liu, Yanan Zheng, Zhengxiao Du, Ming Ding, Yujie Qian, Zhilin Yang, and
  Jie Tang. 2021.
\newblock \href {http://arxiv.org/abs/2103.10385} {Gpt understands, too}.

\bibitem[{Ma et~al.(2019)Ma, Zeng, Zhu, Li, Yang, Yao, Zhou, and
  Shen}]{ma2019end}
Yue Ma, Zengfeng Zeng, Dawei Zhu, Xuan Li, Yiying Yang, Xiaoyuan Yao, Kaijie
  Zhou, and Jianping Shen. 2019.
\newblock \href {https://doi.org/10.48550/ARXIV.1912.09297} {An end-to-end
  dialogue state tracking system with machine reading comprehension and wide \&
  deep classification}.

\bibitem[{Madotto et~al.(2020)Madotto, Liu, Lin, and
  Fung}]{madotto2020language}
Andrea Madotto, Zihan Liu, Zhaojiang Lin, and Pascale Fung. 2020.
\newblock \href {http://arxiv.org/abs/2008.06239} {Language models as few-shot
  learner for task-oriented dialogue systems}.
\newblock \emph{CoRR}, abs/2008.06239.

\bibitem[{Raffel et~al.(2020)Raffel, Shazeer, Roberts, Lee, Narang, Matena,
  Zhou, Li, and Liu}]{raffel2020exploring}
Colin Raffel, Noam Shazeer, Adam Roberts, Katherine Lee, Sharan Narang, Michael
  Matena, Yanqi Zhou, Wei Li, and Peter~J. Liu. 2020.
\newblock \href {http://arxiv.org/abs/1910.10683} {Exploring the limits of
  transfer learning with a unified text-to-text transformer}.

\bibitem[{Rastogi et~al.(2020{\natexlab{a}})Rastogi, Zang, Sunkara, Gupta, and
  Khaitan}]{rastogi2020schema}
Abhinav Rastogi, Xiaoxue Zang, Srinivas Sunkara, Raghav Gupta, and Pranav
  Khaitan. 2020{\natexlab{a}}.
\newblock \href {https://doi.org/10.48550/ARXIV.2002.01359} {Schema-guided
  dialogue state tracking task at dstc8}.

\bibitem[{Rastogi et~al.(2020{\natexlab{b}})Rastogi, Zang, Sunkara, Gupta, and
  Khaitan}]{rastogi2020towards}
Abhinav Rastogi, Xiaoxue Zang, Srinivas Sunkara, Raghav Gupta, and Pranav
  Khaitan. 2020{\natexlab{b}}.
\newblock \href {https://doi.org/10.1609/aaai.v34i05.6394} {Towards scalable
  multi-domain conversational agents: The schema-guided dialogue dataset}.
\newblock \emph{Proceedings of the AAAI Conference on Artificial Intelligence},
  34(05):8689--8696.

\bibitem[{Reynolds and McDonell(2021)}]{reynolds2021prompt}
Laria Reynolds and Kyle McDonell. 2021.
\newblock \href {http://arxiv.org/abs/2102.07350} {Prompt programming for large
  language models: Beyond the few-shot paradigm}.

\bibitem[{Ruan et~al.(2020)Ruan, Ling, Gu, and Liu}]{ruan2020fine}
Yu-Ping Ruan, Zhen-Hua Ling, Jia-Chen Gu, and Quan Liu. 2020.
\newblock \href {https://doi.org/10.48550/ARXIV.2002.00181} {Fine-tuning bert
  for schema-guided zero-shot dialogue state tracking}.

\bibitem[{Wei et~al.(2021)Wei, Bosma, Zhao, Guu, Yu, Lester, Du, Dai, and
  Le}]{wei2021finetuned}
Jason Wei, Maarten Bosma, Vincent~Y. Zhao, Kelvin Guu, Adams~Wei Yu, Brian
  Lester, Nan Du, Andrew~M. Dai, and Quoc~V. Le. 2021.
\newblock \href {http://arxiv.org/abs/2109.01652} {Finetuned language models
  are zero-shot learners}.

\bibitem[{Wu et~al.(2019)Wu, Madotto, Hosseini-Asl, Xiong, Socher, and
  Fung}]{wu2019transferable}
Chien-Sheng Wu, Andrea Madotto, Ehsan Hosseini-Asl, Caiming Xiong, Richard
  Socher, and Pascale Fung. 2019.
\newblock \href {http://arxiv.org/abs/1905.08743} {Transferable multi-domain
  state generator for task-oriented dialogue systems}.

\bibitem[{Zhang et~al.(2021)Zhang, Noroozi, Bakhturina, and
  Ginsburg}]{zhang2021sgdqa}
Yang Zhang, Vahid Noroozi, Evelina Bakhturina, and Boris Ginsburg. 2021.
\newblock \href {http://arxiv.org/abs/2105.08049} {Sgd-qa: Fast schema-guided
  dialogue state tracking for unseen services}.

\bibitem[{Zhao et~al.(2022)Zhao, Gupta, Cao, Yu, Wang, Lee, Rastogi, Shafran,
  and Wu}]{zhao2022description}
Jeffrey Zhao, Raghav Gupta, Yuan Cao, Dian Yu, Mingqiu Wang, Harrison Lee,
  Abhinav Rastogi, Izhak Shafran, and Yonghui Wu. 2022.
\newblock \href {https://doi.org/10.48550/ARXIV.2201.08904} {Description-driven
  task-oriented dialog modeling}.

\bibitem[{Zhao et~al.(2021{\natexlab{a}})Zhao, Mahdieh, Zhang, Cao, and
  Wu}]{zhao2021effective}
Jeffrey Zhao, Mahdis Mahdieh, Ye~Zhang, Yuan Cao, and Yonghui Wu.
  2021{\natexlab{a}}.
\newblock \href {https://doi.org/10.18653/v1/2021.emnlp-main.593} {Effective
  sequence-to-sequence dialogue state tracking}.
\newblock In \emph{Proceedings of the 2021 Conference on Empirical Methods in
  Natural Language Processing}, pages 7486--7493, Online and Punta Cana,
  Dominican Republic. Association for Computational Linguistics.

\bibitem[{Zhao et~al.(2021{\natexlab{b}})Zhao, Wallace, Feng, Klein, and
  Singh}]{zhao2021calibrate}
Tony~Z. Zhao, Eric Wallace, Shi Feng, Dan Klein, and Sameer Singh.
  2021{\natexlab{b}}.
\newblock \href {http://arxiv.org/abs/2102.09690} {Calibrate before use:
  Improving few-shot performance of language models}.

\end{thebibliography}
\bibliographystyle{acl_natbib}

\appendix
\setcounter{table}{0}
\renewcommand{\thetable}{A\arabic{table}}

\section{Prompt Design}
\label{sec:prompt_design_study}

We experimented with various formats for the SDT prompt before arriving at the final format. Below, we list alternative designs that we tried and their impact on JGA, as evaluated on the SGD test set.

\subsection{Categorical value strings vs. multiple choice answers} 
We found that JGA dropped -2\% when we tasked the model with decoding categorical values instead of multiple choice answers - e.g. \texttt{payment\_method=debit card} instead of \texttt{payment\_method=b} (where \texttt{b} is linked to the value \texttt{debit card} in the prompt as described in Section \ref{sec:methodology}). When tasking the model to decode categorical values, it would often decode related yet invalid values, which we counted as false in our evaluation. For example, instead of \texttt{debit card}, the model might decode \texttt{bank balance}.

\subsection{Slot IDs vs. slot names}
When we delexicalized slot names with slot IDs, JGA dropped -5\%. One downside of this approach is that the model lost access to valuable semantic information conveyed by the slot name. Another downside is that the model could not distinguish two slots that had the same value in the prompt. For example, if the prompt was "I would like a pet-friendly hotel room with wifi" and the corresponding slots were \texttt{1=True} (has\_wifi) and \texttt{2=True} (pets\_allowed), it is ambiguous which ID refers to which slot. 

The potential upside of using slot IDs was to remove dependence on the choice of slot name, but this did not succeed for the reasons above.

\subsection{Decoding active slots vs. all slots}
We experimented with training the model to only decode active slots rather than all slots with \texttt{none} values when they were inactive. JGA dropped -0.4\%, which we hypothesized might be a result of greater dissimilarity between the slot-value string in the prompt (which contained all slots by construction) and the target, which only contained a subset of slots.

\subsection{In-line annotations vs. dialogue+slots concatenated}
We hypothesized that bringing the slot annotation in the prompt closer to where it was mentioned in the dialogue might help the model better understand the slot's semantic meaning. We changed the format as follows:

\begin{itemize}
    \itemsep0em
    \item Original: \texttt{[ex] [user] I would like a pet-friendly hotel room with wifi [system] I found ... \textbf{[slot] has\_wifi=True}}
    \item In-line: \texttt{[ex] [user] I would like a pet-friendly hotel room with wifi \textbf{[has\_wifi=True]} [system] I found ...}
\end{itemize}

However, this decreased JGA by more than -20\%. We hypothesized that this was likely due to a mismatch between the prompt's annotations and the target string format, which we did not change.

\section{SDT Model Details}
\label{sec:sdt_model_details}

We used the publicly available T5 checkpoints\footnote{\url{https://github.com/google-research/text-to-text-transfer-transformer/blob/main/released\_checkpoints.md}}. For all experiments, we used a sequence length of 2048, 10\% dropout and a batch size of 16. We used a constant learning rate of $1e-3$ or $1e-4$. All models were trained for 50k steps or until convergence, and each experiment was conducted on either 64 or 128 TPU v3 chips \citep{tpu2017}.

\section{Baseline Models}
\label{sec:baseline_models}
For SGD, we compare against SGP-DST \citep{ruan2020fine}, MRC+WD-DST \citep{ma2019end}, T5-seq \citep{zhao2022description} and T5-ind \citep{lee2021dialogue}.

For MultiWOZ, we compare against TRADE \citep{wu2019transferable}, SUMBT \citep{lee2019sumbt}, TransferQA \citep{lin2021zeroshot}, and T5-seq. Transfer QA is based on T5-large.

\begin{table}[h]
\centering
\setlength{\tabcolsep}{4pt}
\scalebox{0.9}{
    \begin{tabular}{p{2.8cm}ccc}
    \hline
    \textbf{Model} & All & Seen & Unseen \\
    \hline
    SDT-seq + desc & 88.6$\pm$0.9 & 95.7$\pm$0.5 & 86.2$\pm$1.0 \\
    SDT-seq & \textbf{88.8}$\pm$\textbf{0.5} & \textbf{95.8}$\pm$\textbf{0.2} & \textbf{86.4}$\pm$\textbf{0.7} \\
    \hline
    \end{tabular}
}
\caption{We experiment with prompting using both descriptions and demonstrations (SDT-seq + desc) vs. demonstrations-only (SDT-seq) and find that adding descriptions does not improve performance.}
\label{table:desc_and_demo}
\end{table}

\begin{figure}[h]
\centering
\scalebox{0.95}{
\includegraphics[width=1.05\columnwidth]{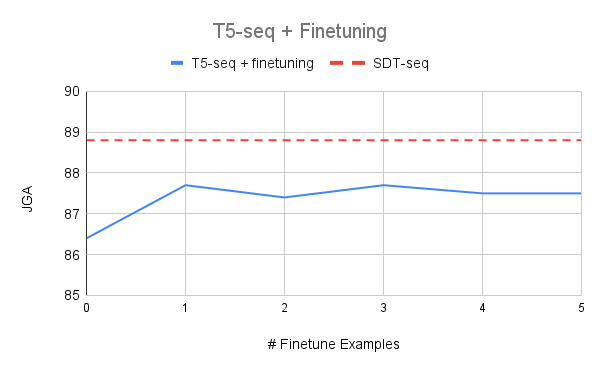}
}
\caption{Results of secondarily finetuning T5-seq with dialogues, to help understand whether prompting or finetuning is more effective. The examples used for finetuning are derived from the set of dialogues used as prompts across the 5 trials of SDT-seq. From this, we observe that prompting with a single dialogue demonstration outperforms few-shot finetuning.}
\label{fig:t5seq_finetune}
\end{figure}

\begin{figure*}
\centering
\includegraphics[width=0.95\textwidth]{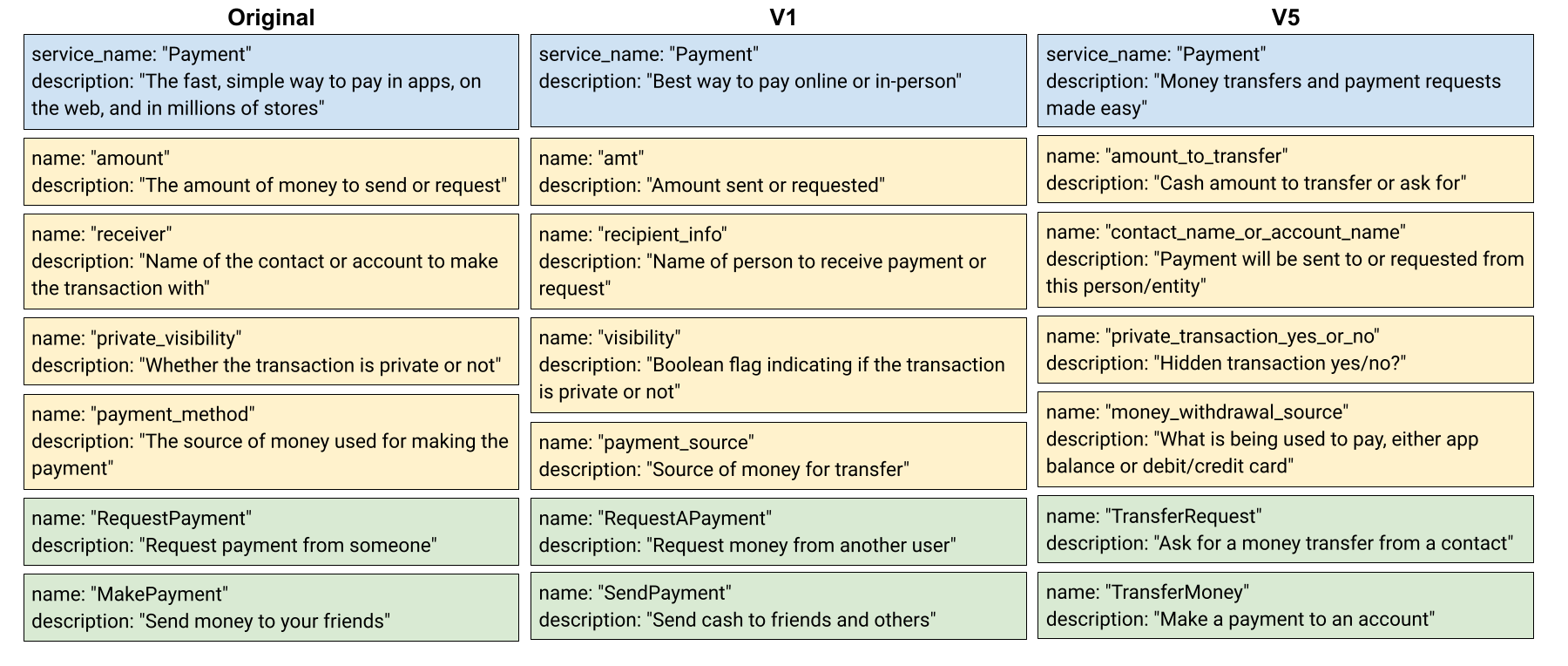}

\caption{The original schema for a Payment service alongside its closest ($v_1$) and farthest ($v_5$) SGD-X variants, as measured by linguistic distance functions. For the SGD-X benchmark, models are trained on the original SGD dataset and evaluated on the test set, where the original test set schemas are replaced by SGD-X variant schemas.}
\label{fig:sgdx_schemas}
\end{figure*}

\end{document}